\documentclass[11pt]{article}

\usepackage[utf8]{inputenc}
\usepackage[T1]{fontenc}
\usepackage[margin=1in]{geometry}
\usepackage{xcolor}
\usepackage{etoolbox}
\usepackage{graphicx}
\usepackage{booktabs}
\usepackage{amsmath}
\usepackage{amssymb}
\usepackage{bm}
\usepackage{caption}
\usepackage{multirow}
\usepackage{array}
\usepackage[round]{natbib}
\usepackage[hidelinks]{hyperref}
\usepackage{url}
\usepackage{pifont}
\newcolumntype{C}[1]{>{\centering\arraybackslash}m{#1}}
\newcommand{\cmark}{\ding{51}}
\newcommand{\xmark}{\ding{55}}

\definecolor{colour1}{RGB}{0,102,204}   
\definecolor{colour2}{RGB}{0,153,76}    
\definecolor{colour3}{RGB}{204,0,0}     
\definecolor{colour4}{RGB}{153,51,255}  
\definecolor{colour5}{RGB}{153,51,255}  

\newtoggle{editingMode}
\settoggle{editingMode}{true}
\iftoggle{editingMode}
{
    \newcounter{noteGLOBALctr} \setcounter{noteGLOBALctr}{1}
    \newcommand{\ps}[1]{\textcolor{colour2}{{{Praveen: }}#1}\par}
    \newcommand{\vk}[1]{\textcolor{colour1}{{{Ville: }}#1}\par}
    \newcommand{\lu}[1]{\textcolor{colour4}{{{Lorenzo: }}#1}\par}
    \newcommand{\todo}[1]{\textcolor{colour3}{{{TODO: }}#1}\par}
}
{
    \newcommand{\ps}[1]{}
    \newcommand{\vk}[1]{}
    \newcommand{\lu}[1]{}
    \newcommand{\todo}[1]{}
}
\title{\vspace{-3em}\textbf{\LARGE ArmnetBench v0.1: Parallel Real-World Evaluation of Manipulation Policies on a Low-Cost Arm Farm}}

\author{
Praveen Selvaraj \quad\quad Lorenzo Uttini \quad\quad Ville Kuosmanen \\[0.5em]
\textit{Armnet} \\
\texttt{armnet.dev}
}

\date{}

\begin{document}
\maketitle

\begin{abstract}
    Real-world evaluation is a bottleneck in developing generalist robot
    manipulation policies. Each rollout requires physical hardware and an operator
    to set up, reset, and score it. We introduce \textbf{ArmnetBench v0.1}, a benchmark
    run on a fleet of low-cost SO-101 cells under light on-site supervision. v0.1
    validates this arm farm end to end and compares 7 policies across 12 tasks with both
    single-arm and bimanual configurations. Each policy is trained or fine-tuned on
    50 demonstrations per task; the benchmark contains 2{,}518 policy rollouts and
    600 reference demonstrations. All 3{,}118 episodes carry a three-way label
    (successful, suboptimal, or failure). Policy rollouts are human-scored, while
    demonstrations are successful by construction. Beyond evaluation, its
    quality-labelled trajectories support downstream learning, from reward and
    predictive world models to policies trained on mixed-quality data. The
    leaderboard is an initial comparison under this shared budget. We release the 3{,}118 core
    episodes in LeRobot~v3.0 and RoboMeter formats.
\end{abstract}

\section{Introduction}
Real-world evaluation increasingly gates progress in generalist robot
manipulation. Assessing competence requires physical hardware, varied trials, and
an operator to arrange, reset, and score each rollout. The resulting cost limits
many studies to a few dozen trials on one robot in one lab, making improvements
hard to distinguish from noise and results hard to compare. Unreported choices
such as camera placement, lighting, object initialisation, checkpoint selection,
and task-specific tuning can further inflate headline success rates and hinder
independent reproduction~\citep{roboarena2025}.

Simulation makes evaluation cheap and reproducible but inherits a sim-to-real gap.
Physical benchmarks avoid that gap, but often trade cost for fewer tasks or
trials, a single embodiment, or coarse binary scoring. ArmnetBench instead reduces the
per-trial cost with parallel low-cost cells. Each adds three cameras, a
Raspberry~Pi, and a networked power plug to the largely 3D-printable
SO-101~\citep{cadene2024lerobot}. The farm runs policy rollouts in isolated
containers under one protocol. This supports broad coverage across tasks,
policies as well as single-arm and bimanual configurations, with every scored rollout
released as data.

v0.1 validates this evaluation substrate and establishes a baseline for iteration.
Its leaderboard compares policies trained or fine-tuned on the same 50
demonstrations per task; it does not estimate each method's capability ceiling.

\medskip
\textbf{Contributions.}
\begin{itemize}
    \item \textbf{The Armnet arm farm}: a managed fleet of low-cost single-arm and
    bimanual SO-101 cells that evaluates policies in parallel under light on-site
    supervision (Section~\ref{sec:farm}).
    \item \textbf{ArmnetBench v0.1}: a 12-task suite and shared-budget evaluation
    protocol for comparing 7 policies across single-arm and bimanual settings,
    with three-way operator scoring
    (Sections~\ref{sec:bench} and~\ref{sec:results}).
    \item \textbf{A released labelled corpus}: 3{,}718 labelled episodes in total,
    comprising 3{,}118 core benchmark episodes in LeRobot~v3.0 and RoboMeter
    formats plus 600 additional rollouts. We also release the exact policy
    checkpoint evaluated for every benchmark task--policy pair
    (Section~\ref{sec:formats}).
\end{itemize}

\section{Related Work}
\label{sec:related}
\textbf{Simulation and task-suite benchmarks.} Simulated suites make manipulation
evaluation cheap and parallel. Meta-World~\citep{yu2020metaworld} covers
multi-task and meta-RL manipulation, LIBERO~\citep{liu2023libero} targets
lifelong-learning transfer, and SIMPLER~\citep{li2024simpler} seeks rankings that
align with real robots. Contact, deformables, and visual realism still transfer
imperfectly, motivating evaluation on physical robots.

\textbf{Reproducible real-world benchmarks.} Physical benchmarks standardise
hardware or scenes. REPLAB~\citep{yang2019replab} provides a reproducible
low-cost cell, SceneReplica~\citep{khargonkar2024scenereplica} replicable scenes,
FurnitureBench~\citep{heo2025furniturebench} long-horizon assembly,
FMB~\citep{luo2025fmb} functional manipulation, and
VLA-REPLICA~\citep{huang2026vlareplica} a replicable SO-101 cell for cross-lab
VLA evaluation. These are primarily blueprints for independently rebuilt setups.

\begin{table}[t]
\centering
\small
\setlength{\tabcolsep}{4pt}
\caption{ArmnetBench versus representative real-world manipulation benchmarks.
Entries are qualitative characterisations of each work's dominant design choice.
\emph{Labelled eval corpus} means a released set of policy \emph{rollouts}, each
tagged with a success/failure (or graded) outcome, as distinct from teleoperated
training demonstrations.}
\label{tab:related}
\begin{tabular}{@{}lllcc@{}}
\toprule
\textbf{Benchmark} & \textbf{Setup} & \textbf{Scoring} & \begin{tabular}[c]{@{}c@{}}\textbf{Human}\\\textbf{in loop}\end{tabular} & \begin{tabular}[c]{@{}c@{}}\textbf{Labelled}\\\textbf{eval corpus}\end{tabular} \\
\midrule
REPLAB \citep{yang2019replab}            & Blueprint cell   & Learned binary    & \xmark & \xmark \\
SceneReplica \citep{khargonkar2024scenereplica} & Blueprint scenes & Binary       & \cmark & \xmark \\
FurnitureBench \citep{heo2025furniturebench}    & Blueprint cell   & Binary $+$ phases & \cmark & \xmark \\
FMB \citep{luo2025fmb}                   & Blueprint cell   & Binary            & \cmark & \xmark \\
AutoEval \citep{autoeval2025}            & Autonomous cell  & Learned binary    & \xmark & \xmark \\
RoboArena \citep{roboarena2025}          & Federated labs   & Preference        & \cmark & \xmark \\
RoboDojo \citep{robodojo2026}            & Sim $+$ remote real & Binary $+$ score & \cmark & \xmark \\
VLA-REPLICA \citep{huang2026vlareplica}  & Blueprint cell   & Binary            & \cmark & \xmark \\
\midrule
\textbf{ArmnetBench (ours)}              & Managed fleet    & Graded 3-way    & \cmark & \cmark \\
\bottomrule
\end{tabular}
\end{table}

\textbf{Auto-evaluation and distributed evaluation.}
AutoEval~\citep{autoeval2025} automates reset and success detection on one robot.
RoboArena~\citep{roboarena2025} pools heterogeneous labs and uses preference
scoring because absolute rates do not compare across sites. RoboDojo~\citep{robodojo2026} combines simulation with a standardised remote
service on bimanual platforms.

\textbf{Low-cost arms and datasets.} ALOHA~\citep{zhao2023aloha} showed that
low-cost hardware can support fine-grained bimanual manipulation. Large-scale
datasets such as DROID~\citep{khazatsky2024droid} collect diverse real-world
interactions, while Open X-Embodiment~\citep{openx2024} aggregates data across
robot embodiments. LeRobot~\citep{cadene2024lerobot} provides open hardware,
software, and dataset tooling around low-cost arms including the SO-100/SO-101.

Together, these lines of work address simulation, reproducible hardware,
automated or distributed evaluation, and scalable data collection. ArmnetBench
combines a managed low-cost fleet with same-cell policy comparison and a released,
graded evaluation corpus (Table~\ref{tab:related}). Whether absolute rates agree
across nominally identical cells remains untested.

\section{The Armnet Arm Farm}
\label{sec:farm}
We collect ArmnetBench v0.1 on the \emph{Armnet} arm farm: a managed fleet of low-cost
single-arm and bimanual SO-101 cells. In v0.1, one co-located workstation builds
and runs policy containers, performs inference and real-time control, and drives
the arms over the local network. A cloud backend schedules jobs and streams logs
and video for operator monitoring.

\subsection{Cells}
\label{sec:cellsetup}
\begin{figure}[t]
\centering
\includegraphics[width=\textwidth]{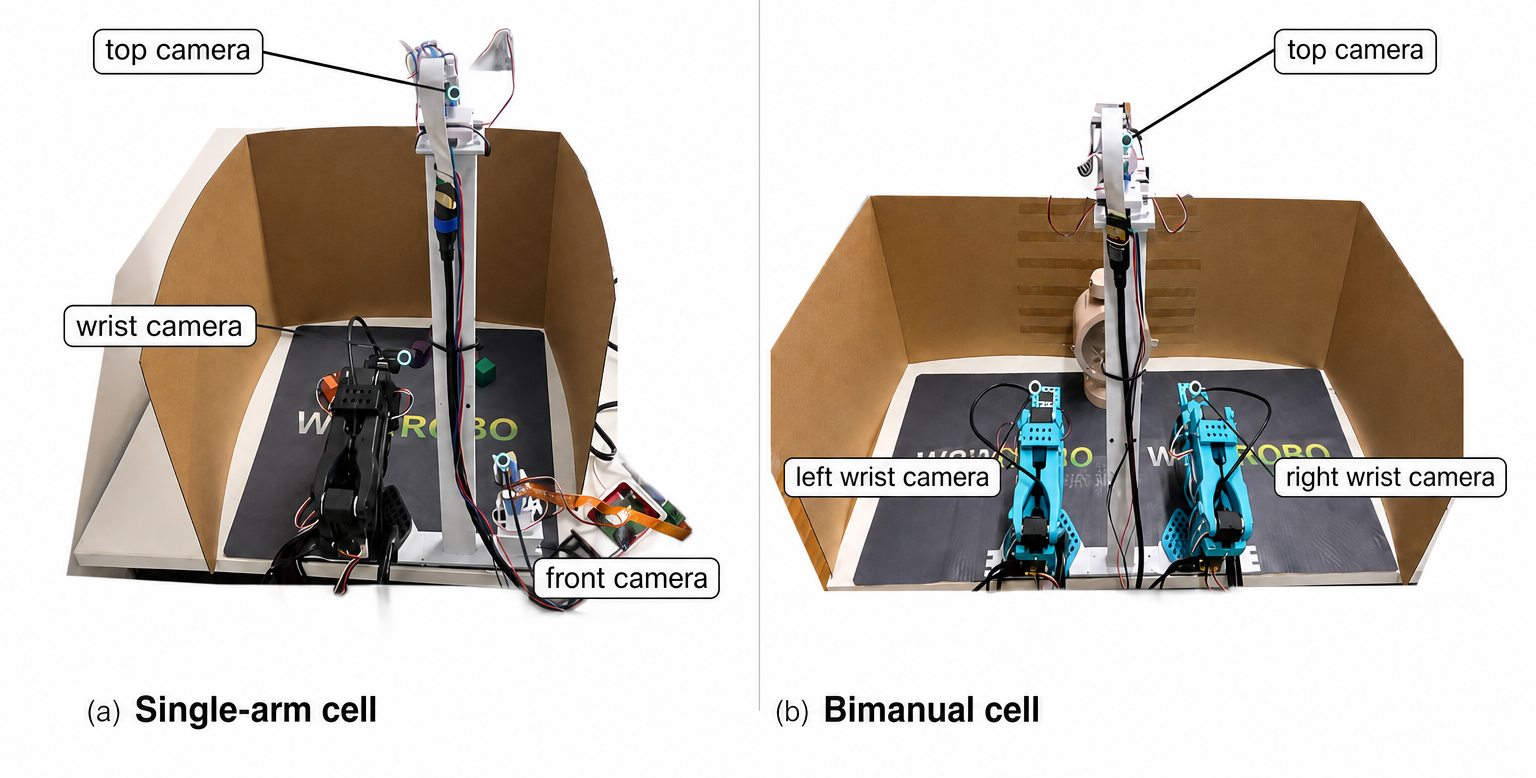}
\caption{Single-arm (left) and bimanual (right) ArmnetBench cells. Each cell
uses a bounded cardboard workspace and three cameras. The single-arm cell uses
\texttt{top}, \texttt{front}, and \texttt{wrist} views; the bimanual cell uses
\texttt{top}, \texttt{left\_wrist}, and \texttt{right\_wrist} views.}
\label{fig:cell}
\end{figure}

Each \emph{cell} contains one or two SO-101 5-DoF follower arms, 3 cameras, and a
Raspberry~Pi~5 edge device in a fixed workspace (Figure~\ref{fig:cell}). The edge
device connects the Feetech serial bus and cameras to the workstation. Observations
are camera views, five arm-joint positions and one gripper position per arm;
actions are target arm-joint and gripper positions. A networked Shelly plug
supports remote power scheduling and recovery.
The nominal cell setup is fixed within each task and object positions are randomised
between rollouts.

\textbf{Enclosure and workspace.} A3 cardboard sheets ($297\times420$\,mm) bound
the workspace and provide a plain backdrop: three sheets for single-arm cells and
four for the wider bimanual cell. One
enclosure used cut-down cardboard, so exact geometry and background differ mildly.

\textbf{Cameras.} Each cell carries three cameras. Single-arm cells use
\texttt{front}, \texttt{top}, and \texttt{wrist} views; bimanual cells use
\texttt{top}, \texttt{left\_wrist}, and \texttt{right\_wrist}. The context cameras
(\texttt{front}/\texttt{top}) are Raspberry~Pi Camera Module~3 units (IMX708) on
Arducam pan/tilt mounts. TheRobotStudio's 3D-printed
\href{https://github.com/TheRobotStudio/SO-ARM100/tree/main/Optional/Overhead_Cam_Mount_Webcam}{overhead-camera
mount} is modified so its original webcam interface accepts the Arducam pan/tilt
unit. Supplementary side brackets attach to this mount and provide clamping points
for the arm bases, fixing top-camera-to-arm geometry and bimanual arm spacing.
Pan/tilt values are stored in the per-cell configuration and held fixed during
evaluation. Innomaker U20CAM 1080p USB wrist cameras mount on the end-effectors.
The \texttt{top} and \texttt{front} cameras sit at approximately 46.9\,cm and
6.3\,cm, respectively. All views use a 16:9 aspect ratio chosen to retain the most scene detail, recording at 20\,fps. Context cameras (\texttt{top}/\texttt{front}) use $1024\times576$ and wrist cameras use
$1280\times720$.

\textbf{Lighting.} v0.1 uses ambient office light with cells away
from windows and all room lights on. Illumination is not calibrated or matched
across cells.

\textbf{Cost.} Table~\ref{tab:bom} gives indicative mid-2026 US retail prices.
The follower price amortises the \$259
\href{https://shop.wowrobo.com/products/so-arm101-diy-kit-assembled-version-1}{SO-ARM101
kit}, including printed parts and 12 servos, across its leader/follower pair.

\begin{table}[t]
\centering
\small
\caption{Approximate per-cell bill of materials (indicative US retail, mid-2026).
Arm, camera, and enclosure quantities differ between the configurations.}
\label{tab:bom}
\begin{tabular}{@{}lC{1.2cm}C{2cm}C{2cm}@{}}
\toprule
\textbf{Component} & \begin{tabular}[c]{@{}c@{}}\textbf{Price}\\\textbf{(USD)}\end{tabular} & \begin{tabular}[c]{@{}c@{}}\textbf{Qty}\\\textbf{single-arm}\end{tabular} & \begin{tabular}[c]{@{}c@{}}\textbf{Qty}\\\textbf{bimanual}\end{tabular} \\
\midrule
SO-101 follower arm & 150 & 1 & 2 \\
Table clamp         & 8   & 2 & 2 \\
Context camera      & 52  & 2 & 1 \\
Wrist camera        & 19  & 1 & 2 \\
Raspberry Pi 5      & 45  & 1 & 1 \\
Shelly smart plug   & 22  & 1 & 1 \\
A3 cardboard sheet  & 1   & 3 & 4 \\
\midrule
\textbf{Total (USD)} & & 359 & 477 \\
\bottomrule
\end{tabular}
\end{table}

\subsection{Running an evaluation}
\begin{figure}[t]
\centering
\includegraphics[width=\textwidth]{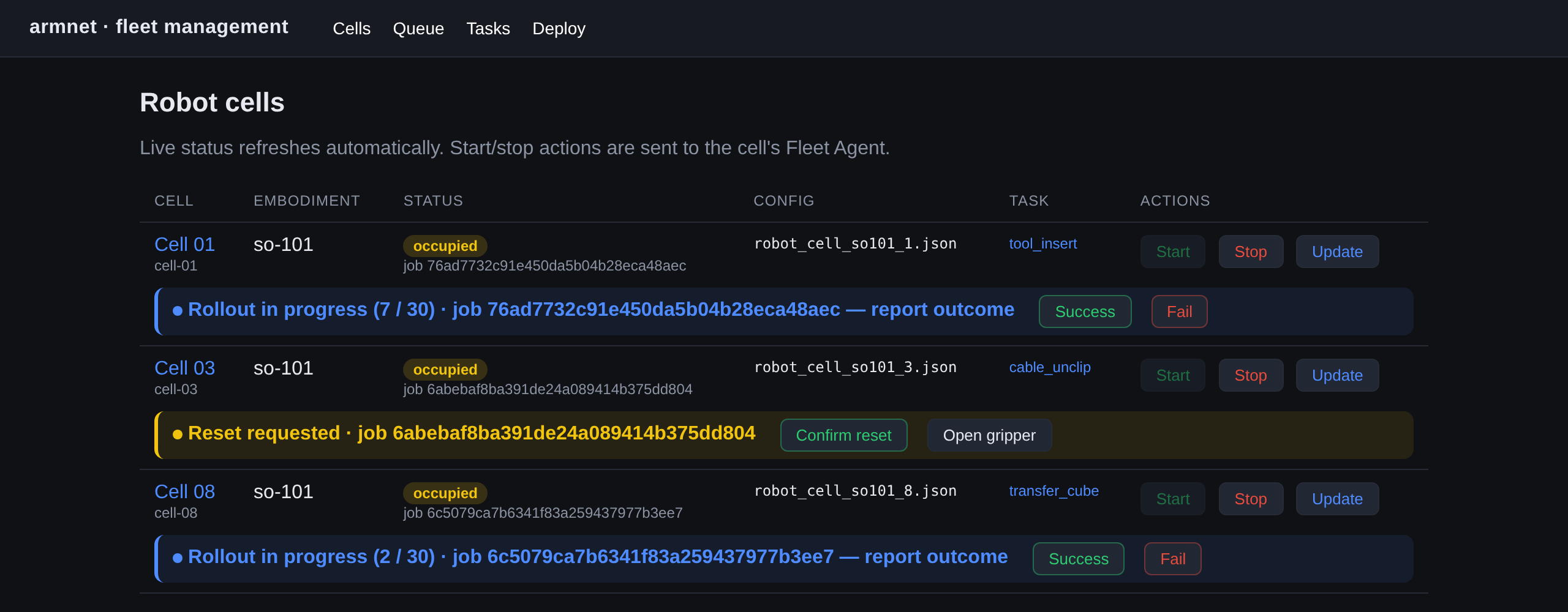}
\caption{The fleet-management control panel used for parallel evaluation. Two
cells are mid-rollout, with progress and scoring controls, while the third is
waiting for the operator to confirm a scene reset. Each row also shows
occupancy, the active job, configuration, task, and cell lifecycle controls.}
\label{fig:panel}
\end{figure}

A user submits a policy image, embodiment, and task. The workstation builds the
image and runs it locally in an isolated container, keeping policy dependencies
separate from cell software. The running policy streams commands to the edge Pi
connected to a matching arm. Logs and video stream to the control panel
(Figure~\ref{fig:panel}), where operators monitor the fleet, reset the scene, and
score each rollout. Opening a cell's detail view shows recent history and
live-streaming logs from its cell and edge processes. Cells execute autonomously
between resets.

\subsection{v0.1 deployment}
\label{sec:sysperf}
We collected v0.1 on 3 cells co-located in a single room: two
single-arm SO-101 cells (\texttt{cell-1}, \texttt{cell-3}) and one bimanual cell
(\texttt{cell-8}). We partitioned tasks across cells so that all seven
policies for a task ran on the \emph{same} cell, holding the nominal setup
constant within that task. The cloud backend could drive geographically
distributed cells, but v0.1 did not exercise this.

Per-cell hardware cost appears in Table~\ref{tab:bom}.
Table~\ref{tab:sysperf} summarises deployment scale and operator workload.
Operators passively monitored all active cells throughout each rollout but were
actively involved only for scene resets and scoring. The active-time figure is
a retrospective estimate rather than a measurement collected during v0.1.

\begin{table}[t]
\centering
\small
\caption{Armnet arm farm deployment statistics for the v0.1 run.}
\label{tab:sysperf}
\begin{tabular}{@{}lC{3cm}@{}}
\toprule
\textbf{Metric} & \textbf{v0.1} \\
\midrule
Cells (single-arm / bimanual) & 2 / 1 \\
Benchmark rollouts collected (total / per cell) & 2{,}520 / 840 \\
Benchmark rollouts removed (manual reset errors) & 2 \\
Cells supervised concurrently per operator & 3 \\
Active operator time per rollout (reset $+$ score) & $\sim$10\,s \\
\bottomrule
\end{tabular}
\end{table}

\section{The ArmnetBench Benchmark}
\label{sec:bench}
ArmnetBench v0.1 turns the arm-farm infrastructure into a shared test bed for
comparing manipulation policies under a common data budget.

\subsection{Tasks}
\begin{figure}[t]
\centering
\includegraphics[width=\textwidth]{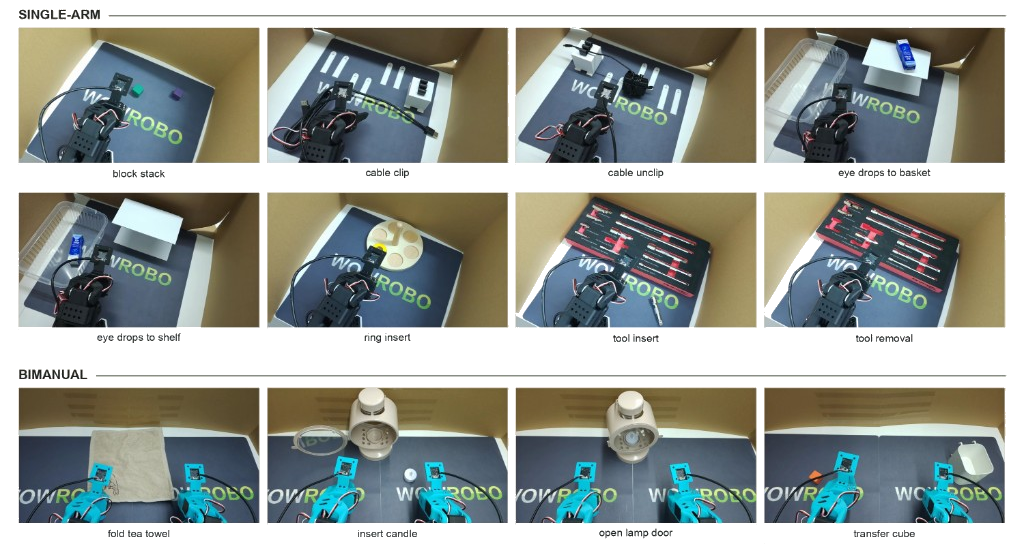}
\caption{The 12 ArmnetBench v0.1 tasks, shown from the \texttt{top} camera at
the start of a human-teleoperated reference demonstration. The upper two rows
show the eight single-arm tasks; the bottom row shows the four bimanual tasks.}
\label{fig:tasks}
\end{figure}

ArmnetBench v0.1 comprises 12 tabletop manipulation tasks, 8 single-arm and 4
bimanual (Table~\ref{tab:tasks}, Figure~\ref{fig:tasks}). The suite spans
stacking, insertion, deformable objects, pick-and-place, and articulated
objects. Every task has 50 human-teleoperated reference demonstrations used to
train or fine-tune each policy.

\begin{table}[t]
\centering
\small
\caption{ArmnetBench v0.1 tasks. Each task has 50 human-teleoperated reference
demonstrations used to train/fine-tune all 7 policies.}
\label{tab:tasks}
\begin{tabular}{@{}llp{7.2cm}@{}}
\toprule
\textbf{Embodiment} & \textbf{Task} & \textbf{Instruction} \\
\midrule
\multirow{8}{*}{Single-arm}
 & \texttt{block\_stack}         & Stack the colourful blocks on top of each other \\
 & \texttt{cable\_clip}          & Push the DisplayPort cable into the cable holder on the white block \\
 & \texttt{cable\_unclip}        & Remove the power cable from the cable holder \\
 & \texttt{eye\_drops\_to\_basket} & Put the eye drops into the basket \\
 & \texttt{eye\_drops\_to\_shelf}  & Put the eye drops on the shelf \\
 & \texttt{ring\_insert}         & Insert the colourful ring into the central wooden peg \\
 & \texttt{tool\_insert}         & Insert the missing tool into the empty slot on the toolbox \\
 & \texttt{tool\_removal}        & Remove the small middle tool from the toolbox \\
\midrule
\multirow{4}{*}{Bimanual}
 & \texttt{fold\_tea\_towel}     & Fold the brown tea towel \\
 & \texttt{insert\_candle}       & Insert the candle inside the lantern and close the door \\
 & \texttt{open\_lamp\_door}     & Hold the lamp still with the left arm and open the door with the right gripper \\
 & \texttt{transfer\_cube}       & Transfer the cube between the arms and drop it into the white basket \\
\bottomrule
\end{tabular}
\end{table}

\subsection{Policies}
We evaluate two specialist imitation policies, ACT~\citep{zhao2023aloha} and
Diffusion Policy~\citep{chi2023diffusion}, and five vision-language-action models:
SmolVLA~\citep{smolvla2025}, $\pi_0$~\citep{black2024pi0},
$\pi_{0.5}$~\citep{pi05_2025}, GR00T~N1.7~\citep{nvidia2025gr00t}, and
MolmoAct~2~\citep{molmoact2_2026}. Each is trained or fine-tuned per task on the
same 50 demonstrations. We release every evaluated task--policy checkpoint.

\subsection{Evaluation protocol and labels}
\label{sec:protocol}
Each task--policy pair targets 30 rollouts. One $\pi_0$ and one $\pi_{0.5}$
episode were removed due to incorrect manual resets, leaving
those pairs with 29. Before each rollout, an operator resets the scene and
randomises object placement within a task-specific range. Initial states are
sampled independently rather than matched across policies. v0.1 did not enforce
standardised per-task wall-clock limits, so rollout termination depended on
operator judgement.
The on-site operator then scores the rollout through the control panel with a
\emph{three-way quality label}:
\begin{itemize}
    \item \textbf{successful} --- the task goal was achieved cleanly
    \item \textbf{suboptimal} --- the goal was achieved, but with a poor-quality finish
    \item \textbf{failure} --- the goal was not achieved
\end{itemize}
The graded scheme can represent partial competence, but operators used
\texttt{suboptimal} conservatively: 89 of 2{,}518 policy rollouts ($3.5\%$) carry
it, so v0.1 is nearly binary. The 50 teleoperated demonstrations per task are
\texttt{successful} by construction and serve as training data and a human
reference. Successful and suboptimal rollouts are trimmed at completion to remove
trailing idle frames; corrupted or unlabelled source episodes are excluded.

In total, v0.1 contains \textbf{3{,}118 labelled episodes}: 600 human-teleoperated
reference episodes ($12 \times 50$) and 2{,}518 policy rollouts
($7 \times 12 \times 30$, minus 2 dropped). Labels comprise 1{,}290 \texttt{successful}, 1{,}739
\texttt{failure}, and 89 \texttt{suboptimal} episodes. Every episode has 3
synchronised camera views at 20\,fps.

These 3{,}118 episodes define the \emph{core benchmark} and all reported
results. The RoboMeter release additionally contains 600 labelled policy
rollouts from extra runs outside the core benchmark. They are released as data
but excluded from the leaderboard and core label totals.

\subsection{Data formats and downstream uses}
\label{sec:formats}
We release the core benchmark in two formats. \textbf{LeRobot~v3.0} stores
synchronised state, action, and packed AV1 video, with sparse terminal reward and done fields
(\texttt{next.reward}, \texttt{next.done}) and episode metadata
(\texttt{success}, \texttt{success\_class}, \texttt{policy\_type},
\texttt{policy\_repo\_id}). The \textbf{RoboMeter} export stores one video per
camera, the \texttt{quality\_label}, and an embedding of the task instruction;
including the auxiliary rollouts, it contains 3{,}718 episodes ($\sim$26 hours).
The two formats preserve the operator-assigned outcomes for downstream analysis.
The quality labels support reward modelling and quality-conditioned policy
training on mixed-quality trajectories. The synchronised observations, actions,
and videos can also be used to train or fine-tune action-conditioned world
models and video-generation modules within robot policies.

\section{Results}
\label{sec:results}

All policies share the 50-demonstration training budget. We report
\emph{strict success rate} over $n$ scored rollouts and the more lenient
\emph{success$+$suboptimal} rate, which also counts poor-quality completion.

\begin{table}[t]
\centering
\small
\caption{ArmnetBench v0.1 overall leaderboard, pooled across all 12 tasks.
Strict success counts only \texttt{successful} rollouts; the last column also
credits \texttt{suboptimal} completions. $n$ is the number of scored rollouts.}
\label{tab:leaderboard}
\begin{tabular}{@{}lC{1cm}C{2cm}C{2.4cm}@{}}
\toprule
\textbf{Policy} & $n$ & \textbf{Success (\%)} & \textbf{Succ.$+$Subopt.\ (\%)} \\
\midrule
$\pi_{0.5}$    & 359 & 47.6 & 51.5 \\
$\pi_0$        & 359 & 35.1 & 40.4 \\
GR00T~N1.7     & 360 & 29.4 & 33.1 \\
Diffusion      & 360 & 26.7 & 29.7 \\
ACT            & 360 & 19.2 & 21.1 \\
MolmoAct~2     & 360 & 18.9 & 21.7 \\
SmolVLA        & 360 & 15.0 & 19.2 \\
\bottomrule
\end{tabular}
\end{table}

$\pi_{0.5}$ leads both embodiments ($45.4\%$ single-arm, $52.1\%$ bimanual), with an overall strict success rate of $47.6\%$. The middle reshuffles on bimanual tasks -- Diffusion Policy rises to second ($35.8\%$), while ACT falls to $2.5\%$. Relative standings are therefore
embodiment-dependent and should not be extrapolated from single-arm results.
Table~\ref{tab:pertask} reports strict success for every (task, policy) pair.
No policy succeeds on the contact-rich \texttt{cable\_clip}, whereas the best
task--policy pairs reach 60--86\%.

\begin{table}[t]
\centering
\scriptsize
\setlength{\tabcolsep}{4pt}
\caption{Per-task strict success rate (\%); entries generally aggregate 30
rollouts, with two at 29 after data cleaning. Best per row in \textbf{bold}.
Single-arm tasks above the rule, bimanual below.}
\label{tab:pertask}
\setlength{\extrarowheight}{2pt}
\begin{tabular}{@{}l*{7}{C{1cm}}@{}}
\toprule
\begin{tabular}[c]{@{}l@{}}\textbf{Task}\end{tabular}
& \multicolumn{1}{c}{\begin{tabular}[c]{@{}c@{}}\textbf{ACT}\end{tabular}}
& \multicolumn{1}{c}{\begin{tabular}[c]{@{}c@{}}\textbf{Diffusion}\\\textbf{Policy}\end{tabular}}
& \multicolumn{1}{c}{\begin{tabular}[c]{@{}c@{}}\textbf{SmolVLA}\end{tabular}}
& \multicolumn{1}{c}{\begin{tabular}[c]{@{}c@{}}$\bm{\pi_0}$\end{tabular}}
& \multicolumn{1}{c}{\begin{tabular}[c]{@{}c@{}}$\bm{\pi_{0.5}}$\end{tabular}}
& \multicolumn{1}{c}{\begin{tabular}[c]{@{}c@{}}\textbf{GR00T}\\\textbf{N1.7}\end{tabular}}
& \multicolumn{1}{c}{\begin{tabular}[c]{@{}c@{}}\textbf{MolmoAct~2}\end{tabular}} \\
\midrule
\texttt{block\_stack}          &  0 &  0 &  7 & 33 & \textbf{43} & 27 &  7 \\
\texttt{ring\_insert}          & 50 & 23 & 33 & 17 & 47 & \textbf{60} & 13 \\
\texttt{tool\_insert}          & 20 & 23 &  7 & 40 & \textbf{53} & 17 &  3 \\
\texttt{tool\_removal}         & 27 & \textbf{63} &  3 & 10 & 13 & 30 &  0 \\
\texttt{cable\_clip}           &  0 &  0 &  0 &  0 &  0 &  0 &  0 \\
\texttt{cable\_unclip}         & 60 &  7 & 37 & 47 & \textbf{70} & 33 & 23 \\
\texttt{eye\_drops\_to\_basket}& 63 & 43 & 23 & \textbf{70} & 67 & 67 & 63 \\
\texttt{eye\_drops\_to\_shelf} &  0 & 17 & 10 & \textbf{76} & 70 & 43 & 47 \\
\midrule
\texttt{fold\_tea\_towel}      & 10 & 50 & 17 & 50 & \textbf{70} & 13 & 20 \\
\texttt{insert\_candle}        &  0 &  7 &  7 & 27 & \textbf{30} &  3 &  3 \\
\texttt{open\_lamp\_door}      &  0 & \textbf{73} & 30 &  7 & 23 & 13 & 33 \\
\texttt{transfer\_cube}        &  0 & 13 &  7 & 47 & \textbf{86} & 47 & 13 \\
\bottomrule
\end{tabular}
\end{table}

\section{Discussion and Limitations}
\label{sec:discussion}
\textbf{Reproducibility.} v0.1 has three levels. (i)~\emph{Execution
reproducibility} is strong: released container images and task--policy checkpoints
reproduce the policy stack. (ii)~\emph{Within-cell scene consistency} is moderate:
rigid mounts and recorded pan/tilt settings fix camera-to-arm geometry, but ambient
lighting varies.
(iii)~\emph{Cross-cell physical reproducibility} remains unanswered because each task
ran on one cell.

\medskip
\textbf{Policy caveats.} Absolute rates depend on our training, fine-tuning, and
integration recipes and should be read with the per-task breakdown in
Section~\ref{sec:results}.

\medskip
\noindent\textbf{Uncontrolled factors.}
\begin{itemize}
    \item \textbf{Manual reset errors.} Two episodes were excluded during data
    cleaning after incorrect scene resets, one from a $\pi_0$ pair and one from
    a $\pi_{0.5}$ pair, leaving those pairs with 29 rollouts.
    \item \textbf{Unmeasured initial-state variation.} Operators randomised object
    placement within task-specific ranges but did not record positions. This
    prevents analysis of spatial success patterns. Unknown training distributions
    may also preclude strict in- versus out-of-distribution classification.
    \item \textbf{Unstandardised rollout duration.} v0.1 did not enforce fixed
    per-task wall-clock limits. Operator-dependent stopping decisions could
    affect failure rates and reduce comparability across task--policy pairs.
    \item \textbf{Object deterioration.} Some task objects wore down over the
    evaluation, including the eye-drops carton and cable-task velcro, potentially
    changing task difficulty.
    \item \textbf{Physical cell changes.} On \texttt{cell-3} the \texttt{front}
    camera was misaligned for every policy except MolmoAct~2, although the full
    scene remained visible from \texttt{top}. The bimanual
    \texttt{right\_wrist} view was slightly blurry for every policy except
    MolmoAct~2. Either issue could depress affected success rates.
\end{itemize}

\section{Future Work}
\label{sec:futurework}
The primary future objective is reducing operator time per rollout. Automated scene
reset would remove the main manual step and make initial states repeatable;
initial-position logging would verify each reset, and a reward-model labeller
would reduce manual scoring. Together, these changes should let one operator
supervise more cells. Secondary goals are a broader task suite, per-cell
lightboxes, camera-framing checks, and a shared
task--policy slice across cells to measure cross-cell reproducibility.

\section{Conclusion}
ArmnetBench demonstrates real-world policy evaluation on low-cost parallel
SO-101 cells under light on-site supervision. v0.1 validates the system end to
end and produces a graded rollout corpus. Its quality-labelled trajectories also
support training and fine-tuning a range of downstream robot learning models.
Its leaderboard compares evaluated implementations under one training budget.
We release the core benchmark in LeRobot~v3.0 and RoboMeter formats, together
with every benchmark task--policy checkpoint, on the HuggingFace
Hub.\footnote{\url{https://huggingface.co/collections/armnet/armnetbench-v01}}

\bibliographystyle{plainnat}
\bibliography{references}

\end{document}